\pdfoutput=1

\documentclass[11pt]{article}

\usepackage[preprint]{acl}

\usepackage{times}
\usepackage{latexsym}

\usepackage[T1]{fontenc}

\usepackage[utf8]{inputenc}

\usepackage{microtype}

\usepackage{inconsolata}

\usepackage{graphicx,multirow,array}
\graphicspath{{images/}}
\usepackage{hhline}
\usepackage{subfigure}
\usepackage{booktabs} 
\newcommand\Tstrut{\rule{0pt}{2.6ex}}         
 
\usepackage{amsmath}
\usepackage{amssymb}
\usepackage{mathtools}
\usepackage{amsthm}

\title{Exploring the Impact of a Transformer's Latent Space Geometry on Downstream Task Performance}

\author{
  \textbf{Anna C. Marbut\textsuperscript{1,\dag}},
  \textbf{John W. Chandler\textsuperscript{2}},
  \textbf{Travis J. Wheeler\textsuperscript{3}}
\\
\\
  \textsuperscript{1}Interdisciplinary Studies,  University of Montana, Missoula, MT, USA\\
  \textsuperscript{2}College of Business, University of Montana, Missoula, MT, USA\\
  \textsuperscript{3}R. Ken Coit College of Pharmacy, University of Arizona, Tucson, AZ, USA
\\
  \small{
    \textbf{\dag Correspondence:} \href{mailto:email@domain}{anna.marbut@umontana.edu}
  }
}

\begin{document}
\maketitle
\begin{abstract}
It is generally thought that transformer-based large language models benefit from pre-training by learning generic linguistic knowledge that can be focused on a specific task during fine-tuning. However, we propose that much of the benefit from pre-training may be captured by geometric characteristics of the latent space representations, divorced from any specific linguistic knowledge. In this work we explore the relationship between GLUE benchmarking task performance and a variety of measures applied to the latent space resulting from BERT-type contextual language models. We find that there is a strong linear relationship between a measure of quantized cell density and average GLUE performance and that these measures may be predictive of otherwise surprising GLUE performance for several non-standard BERT-type models from the literature \cite{alajrami2022does, sinha2021masked, zhang2021general}. These results may be suggestive of a strategy for decreasing pre-training requirements, wherein model initialization can be informed by the geometric characteristics of the model's latent space.

\end{abstract}

\section{Introduction}

Transformer-based large language models, such as BERT \cite{devlin2018bert} and the GPT models \cite{radford2018improving}, have outperformed other language models on linguistic benchmarking tasks time and again. These models largely rely on a pre-training and fine-tuning training protocol, where the base model is trained with a very large corpus on a generic language task, such as masked language modeling, and then the weights are updated in a second round of training with a more specific task, such as text similarity or entailment. The assumption motivating this process is that the pre-training task provides the model with general linguistic information, which allows it to better learn a specific linguistic task during fine-tuning. 

However, this assumption has been challenged in several recent works in which the BERT model training process is altered such that we would expect the model to perform poorly on benchmarking tasks. Examples of these pathological changes include using a non-linguistic training task or shuffling the model's input embeddings \cite{alajrami2022does, sinha2021masked, zhang2021general}. If the outstanding benchmarking performance of the standard models relied on the presence of general linguistic knowledge in the underlying model, then these alterations should reduce that performance to a level similar to an untrained model. However, these works consistently find benchmarking performance that is closer to that of a standard model than to that of an untrained model.

We propose that at least some of the benchmarking performance seen with this pre-training/fine-tuning protocol can be attributed to geometric characteristics of the latent space that result from the transformer architecture, divorced from any generic linguistic knowledge. Many common downstream tasks for language models, such as classification or paraphrase, use measures of similarity or separability in their loss functions. These measures rely on the way that latent representations are organized in this space, so it stands to reason that there could be a measurable characteristic of this organization that would be advantageous for learning these fine-tuning tasks. If the latent space is already organized with data placed in various high-dimensional nooks and crannies, it may be easier for a powerful transformer model to learn how to access the data appropriately during fine-tuning, even if the organization scheme is non-linguistic.

Here we explore a variety of measures that could be used to describe the geometry of a high-dimensional contextual language model latent space, and we examine the relationships between these measures and the GLUE benchmarking task performance on a range of BERT-type contextual language models. We introduce a quantized cell density measure that appears to have a strong linear relationship with GLUE performance when applied to latent space representations. We also find that this measure can be used to predict the surprising GLUE performance of many of the non-standard models presented in the aforementioned works, though the performance of a few of these models remains unexplained.

\section{Related Work}

When \citet{vaswani2017attention} introduced the transformer model, it was as an end-to-end autoencoder that relied on the attention mechanism to provide contextual information for sequential data modeling. That architecture was adapted to be used in large language models by isolating either the encoder component (as in the BERT models from \citealt{devlin2018bert}) or the decoder component (as in the GPT models from \citealt{radford2018improving}) and training with a very large dataset on a generalized linguistic task, such as masked language modeling. These models are intended to be fine-tuned on specific downstream tasks, generally with a much smaller, labeled dataset, by adding an additional layer and jointly training the pre-trained and new layer weights on the new task.

It has been shown that this pre-training/fine-tuning procedure produces improved performance on downstream tasks, both on transformer models \cite{devlin2018bert, radford2018improving} and on more traditional neural models \cite{dai2015semi, howard2018universal}. The advantage of this dynamic is generally understood to fall under the theory of transfer learning, whereby the model learns general information during the pre-training task, and this can then be specialized during fine-tuning. This means that many fewer parameters need to be trained from scratch on the fine-tuning tasks, allowing for large models to be effectively applied to these tasks without drastically overfitting to the (often small) supervised training datasets.

Benchmarking datasets provide the research community with the ability to compare performance on the same exact tasks, and generally aim to encapsulate some knowledge or skill that would be useful for the type of model being evaluated. The pre-training/fine-tuning process lends itself nicely to benchmarking measures, as the limited-size benchmarking dataset can be used to fine-tune a very large model. The GLUE benchmarking suite is a collection of tasks for language models, and each of the nine datasets is focused on evaluating a linguistic skill, from sentiment to paraphrase to inference \cite{wang2018glue}. Although other language model benchmarks have been proposed, GLUE remains a key set of measures for evaluating language model performance, and, at the time of writing this, nine of the ten top performing models are transformer-based models, and all ten follow the pre-training/fine-tuning paradigm \cite{GLUEBenchmark}.

However, several recent works bring into question the source of this impressive benchmarking performance. \citet{sinha2021masked} showed that RoBERTa models perform comparably to the standard model when trained with randomly shuffled word order. \citet{zhang2021general} found that using English-trained input embeddings with BERT models trained in another language has surprisingly little effect on the GLUE performance, as does shuffling the indices of trained input embeddings. \citet{alajrami2022does} found that RoBERTa models with non-linguistic pre-training tasks, described in Section \ref{sec:alt_models}, still perform surprisingly well on GLUE tasks. 

The unexpected performance on the GLUE tasks found in these papers suggests that the benefit these models get from pre-training may not be due to generalized linguistic knowledge. \citet{sinha2021masked} suggest that the models may simply be learning complicated distributional co-occurrence statistics, while \citet{zhang2021general} suggest that the linguistic information may actually be contained largely within the trained input embeddings. We instead propose that the benefit of the pre-training process may be related to the geometric characteristics of a latent space resulting from training a model with the transformer architecture, divorced from any specific content knowledge.

Several works have shown that increasing the isotropy of a language model's latent space, or causing the resulting representations to be more uniformly spread in all directions, may increase performance on benchmarking tasks \cite{Mu2017-ck, Liang2021-aq, Kaneko2020-fq, Gao2019-zg}. The motivation behind this work rests on the finding that the latent spaces of these language models are highly anisotropic, with all of the data concentrated in a narrow ``cone'', rather than spread throughout the available space and dimensions \cite{Mimno2017-mj, Mu2017-ck, Ethayarajh2019-fl}. By adjusting the models to have more isotropic latent spaces through training or retrofitting methods, these works have attempted to cause the models to use more of the available space, producing some improvement on selected benchmarking tasks. However, there is a growing body of evidence that an isotropic latent space may not be as beneficial as initially thought. \citet{ding2021isotropy} suggest that the benefits found in earlier work may be isolated to static word embedding models, like Word2Vec and GloVe. \citet{rudman2023stable} find that increasing isotropy is detrimental to contextual model performance when measured with a more reliable measure than used in previous work \cite{marbut2023reliable, rudman2021isoscore}. \citet{cai2020isotropy} and \citet{rajaee2021cluster} find that increasing overall isotropy is not beneficial in contextual latent space models, but that increasing isotropy within local clusters may improve model performance.

Similarly, it has been found that applying quantization methods during model training or fine-tuning may improve benchmarking performance as well \cite{liao2020embedding, sablayrolles2018spreading, hu2022empirical}. Quantization is a compression method in which high-dimensional data are typically clustered within some number of subspaces that can be recombined to recover the full-dimensional data with minimal loss \cite{gholami2022survey}. Minimizing this loss during some part of the model training could result in a latent space that is more easily separable, which may prepare data to be more easily partitioned during fine-tuning on downstream tasks.

\section{Methods}

To explore the question of how the geometric characteristics of a contextual language model's latent space may be related to benchmarking performance, we produced a series of 170 BERT-type models with varying levels of noise added to the pre-trained weights of all the encoder layers, leaving the pre-trained input embeddings un-noised. We then examined the relationship between the GLUE benchmark performance of these models and the values of measures intended to approximate characteristics of data spread and separability in the models' latent spaces. Finally, we explored the possibility of using these measures to predict the surprising GLUE performance of non-standard models from the literature.

\subsection{BERT Models with Injected Noise}

Starting with a pre-trained BERT-type model, we gradually inject noise into all encoder layer weights as shown in Equation \ref{eqn:noise_add}. Here, $W_O$ is the complete pre-trained network of encoder layer weights, noise vector $\epsilon \sim N(0,1)$ is a set of random weight modifiers with the same dimensions, and $\alpha_i$ is a hyperparameter describing the fractional amount of noise in model $i$.
\vspace{-0.05in}
\begin{equation}\label{eqn:noise_add}
    W_i = (1-\alpha_i)W_O +\alpha_i \epsilon
\end{equation}
\noindent
This process resulted in a dataset of 125 perturbed BERT-small models (\citealt{turc2019well}, 5 transformer blocks and 29.1M total parameters) and 8 BERT-base models (\citealt{devlin2018bert}, 12 transformer blocks and 110M total parameters), as well as 38 perturbed RoBERTa-base models (\citealt{liu2019roberta}, 12 transformer blocks and 125M total parameters) .

After perturbing the pre-trained weights, we fine-tuned the models on the GLUE datasets using the script and default parameters from the Transformers library \cite{wolf-etal-2020-transformers}. Following \citet{devlin2018bert}, we excluded the WNLI inference task from our results due to known issues with the dataset \cite{GLUEBenchmarkFAQ}.

\begin{figure}[b]
\vspace{-0.25in}
\begin{center}
\centerline{\includegraphics[width = 0.95\columnwidth]{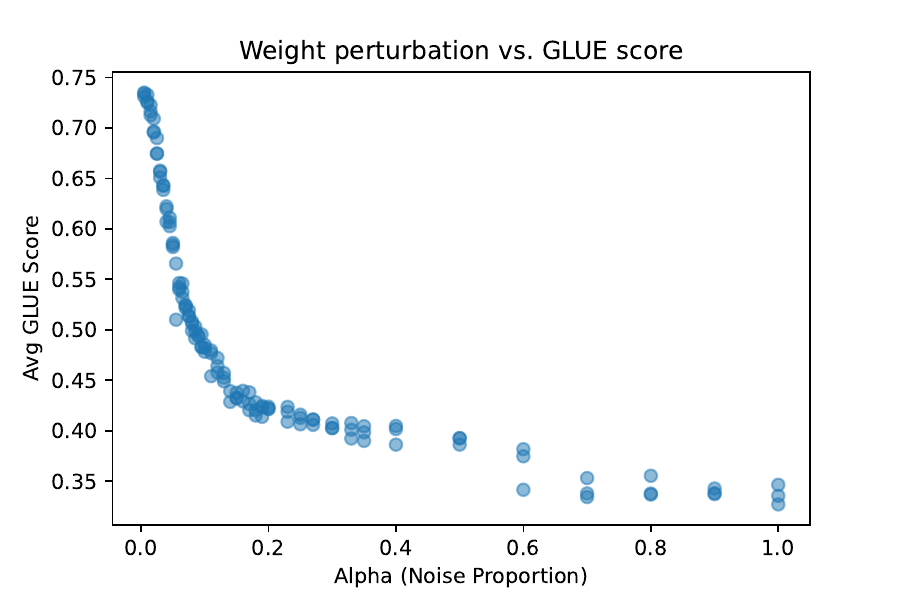}}
\vskip -0.1in
\caption{Relationship between weight perturbation (Equation \ref{eqn:noise_add}) on pre-trained BERT-small and the resulting average GLUE score.}
\label{fig:alpha_glue}
\end{center}
\vskip -0.4in
\end{figure}

As seen in Figure \ref{fig:alpha_glue}, as $\alpha$ increases, or as the weights are transitioned from their trained values to random noise, the GLUE score decreases exponentially. We focused most of our experimentation on lower $\alpha$ levels to capture as much signal as possible during that rapid decrease in GLUE performance, with $\alpha\leq0.1$ for 70 of our 125 BERT-small models. 

\subsection{Latent Space Sampling}
 Much of the existing work that describes the distribution of representations in natural language model latent spaces is centered on static embeddings \cite{Mimno2017-mj, Mu2017-ck, Gong2018-ko, Gao2019-zg}. Studying the shape of a contextual model's latent space is complicated by the fact that the latent representation of each token in the vocabulary will depend on the context in which it is seen, presenting a practically infinite number of representations of each individual token. Many works exploring these distributions largely focus on very local statistics or the relationships between the representations of specific tokens \cite{Ethayarajh2019-fl, bihani2021low, valeriani2024geometry, merchant2020happens}, and although both \citet{cai2020isotropy} and \citet{ferner2021isotropic} are focused on an examination of entire latent spaces, to the best of our knowledge, there is not currently a standard way to produce a representative sample of these spaces as a whole.  

\begin{figure}
\begin{center}
\centerline{\includegraphics[width = \columnwidth]{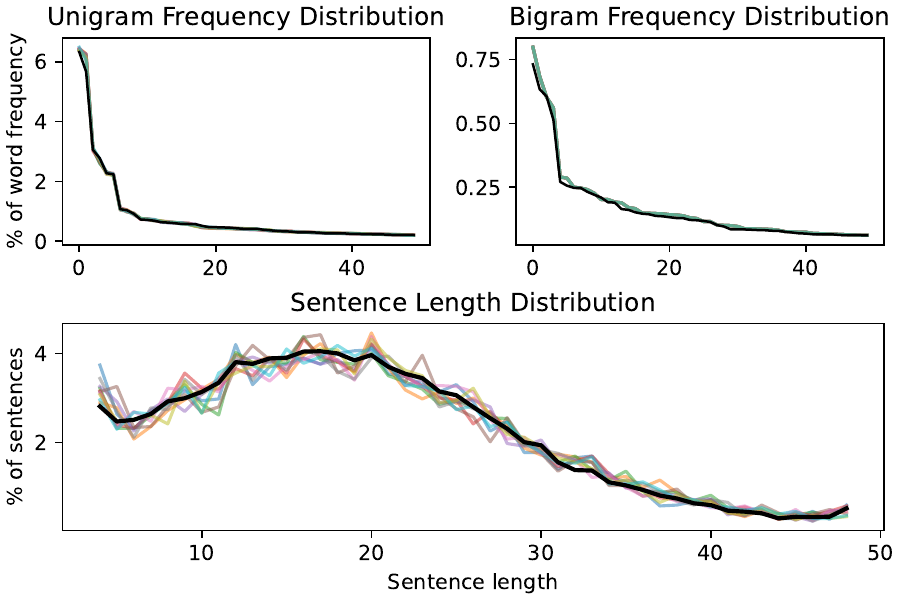}}
\vskip -0.1in
\caption{Sample Sequence unigram, bigram, and sentence length frequencies compared to the complete combined PennTreeBank and WikiText2 datasets.}
\label{fig:sample_dist_compare}
\end{center}
\vskip -0.4in
\end{figure}

As such, we opted to sample 5K sequences (roughly 250K tokens) from the complete Penn Tree Bank (PTB) \cite{marcus1993building} and WikiText2 \cite{merity2016pointer} datasets, including train, validation, and test splits\footnote{We use the scikit-learn \cite{scikit-learn} unigram tokenizer to only include unique sampled sequences between three and fifty tokens, as a surprising number of the sequences from these datasets are duplicates or empty strings.}.  To ensure that we produced a reasonably representative latent space sample, we compared the frequency distributions of unigrams, bigrams, and sentence lengths in ten separate samples to the distributions over the entire combined dataset. As seen in Figure \ref{fig:sample_dist_compare}, the sample distributions closely follow the whole dataset distribution in all three plots. 

\subsection{Measures} \label{sec:metrics}

Although distributional statistics are well understood and established in low dimensions, many common measures, and our intuitions about them, fail as the number of dimensions grows above about seven \cite{marbut2023reliable}. As such, we identified a variety of measures that may approximate how data  are organized in a high dimensional latent space, in particular considering whether the space is used evenly (data spread) and whether the data are organized in a clustered manner.

\subsubsection{Data Spread}

As a simple first pass, we apply a measure of data spread, which can quantify whether the data are distributed evenly in all directions and along each dimension. The concept of data spread may not provide  much detail about the data distribution (it does not offer any insight into the nature of an uneven/irregular distribution), but it is ultimately simpler and more interpretable than our quantization-based measures (see below). Here, we describe one measure of overall data spread; two others are described in the Appendix.

\paragraph{Eigenvalue Early Enrichment (EEE)} \quad
Following \citet{marbut2023reliable}, we consider the cumulative sum of the eigenvalues of a distribution. These eigenvalues, often referred to as the explained variance of the principle components, will all be equal when data is spread evenly. Conversely, almost all of the explained variance will be concentrated in the first few components in a very uneven distribution. Equation~\ref{eq:EEE} calculates the area between the cumulative sum curve, $X_{EEE}$, and the linear sum for a perfectly even distribution, $Y_{ref}$, as a proportion of the total space available above that linear sum line (with $v$ as the total variance in the $d$-dimensional distribution). We expect the EEE of a distribution to decrease as data becomes more evenly spread.
\begin{align}
EEE &= \frac{AUC(X_{EEE}-Y_{ref})}{\frac{1}{2}dv} \label{eq:EEE}
\end{align}

\subsubsection{Quantization}

Another way to assess the organization of a data distribution is to quantify how easily it can be broken into distinct and separable clusters. 
Quantization can be seen as a clustering method that is particularly attractive when working with large, high-dimensional datasets. Typically, in quantization, a high-dimensional data distribution is broken into $m$ subspaces and then the data are clustered within each subspace to create a compressed representation of the original data. 

We take advantage of the speed-optimized quantization methods available through the FAISS library \cite{douze2024faiss}. In particular, we examine measures based on the results of an additive quantizer (LSQ++ from \citealt{martinez2018lsq++}) which iteratively adjusts the assignment of $m=4$ subspaces and $k=256$ cluster centroids to minimize data loss. We describe a few measures based on how data are distributed among or within the quantized bins, which aim to capture the extent to which the quantized points are clustered. Results from several conceptually redundant measures and the full results of using a product quantizer \cite{jegou2010product} can be found in Appendix \ref{app:full_results}.

While calculating measures in the quantized latent space may help to quantify how separable or clustered the space is, we note that the process of quantization is, itself, an imperfect proxy for data separability. 
In particular, as with traditional clustering methods, hyperparameters such as cluster/bin count can have a dramatic effect on the result, and there is not a single standard method for identifying ideal hyperparamters. Here, we rely on a default bin count of 256, and note the potential benefit of further exploration of this and other hyperparameters.

\paragraph{Point Patchiness (PP)} \quad
The distribution of how points are assigned to centroids in a quantized space, or the observed density of those quantized bins, may tell us how uniformly data are distributed. For a uniform latent space, we would expect the cell density to be the same for all centroids, whereas an irregular space would have greater variation in cell density between clusters.

We borrow a measure from ecology~\cite{wade201850} and compute the patchiness index over the quantized latent spaces. This measure is a variance-adjusted average cell density that approximates the cell density as seen from the perspective of an individual datapoint. In Equation \ref{eq:patch}, $m$ is the average cell density and $V$ is the variance of cell densities .
\begin{align}
    PP &= \frac{m+(\frac{V}{m}-1)}{m} \label{eq:patch}
\end{align}

The patchiness index is typically applied over equal-sized bins by design, a state not expected to be the case for our quantized clusters. However, our empirical experiments in weighting the cell densities by their volume (approximated as n-dimensional spheres with a radius equal to the distance between the centroid and its furthest assigned point) resulted in almost all of the densities becoming zero. This should not be surprising given the unintuitive nature of high-dimensional geometry, wherein the volume of a hypersphere with a fixed radius quickly goes to zero as the dimensions increase above seven \cite{marbut2023reliable}. Due to this oddity, we do not include any cell weighting in our patchiness calculation.

\paragraph{Reconstruction Skew (RS)} \quad
In a uniform distribution, each quantized cell will be about the same size, and there will be limited variation in the distance between an original datapoint and its reconstructed counterpart (the error magnitude, Equation \ref{eqn:error_mag}). In a very clustered distribution, some bins will be very small (where the data are clustered) while others will be large (where the data are more sparse). We expect the error magnitude to be small for data in tight clusters and to be large for data in sparsely populated bins. 

We calculate the skew of this distribution \cite{kokoska2000crc}, shown in Equation \ref{eqn:error_skew}, where $\overline{EM}$ is the mean and $s(EM)$ is the standard deviation. Here we would expect the skew of a uniform distribution to be near zero, whereas an irregular, more clustered distribution would result in a left (negative) skew.
\begin{align}
    EM(x_i) &= \frac{\parallel x_i - x'_i \parallel}{\max\limits_{\forall x}\parallel x - x' \parallel} \label{eqn:error_mag} \\
    RS &= \frac{n\sum_{i=1}^{n}\left(\frac{EM(x_i)-\overline{EM}}{s(EM)}\right)^3}{(n-1)(n-2)} \label{eqn:error_skew}
\end{align}

\paragraph{Centroid Distribution (CD\textsubscript{var})} \quad 
We also consider how the centroids themselves are distributed within the $m$ subspaces. In a uniform latent space, we would expect the centroids to be evenly spaced and have very similar nearest-neighbor distances, but in a more irregular space we would expect to see greater variation in these distances. For each centroid we calculate the normalized distance to the nearest centroid ($\rho_i$) and report the variance of these distances as a measure over the combined subspaces (Equation \ref{eqn:cdv}). We use the FAISS L2 nearest neighbor implementation to efficiently find the approximate nearest neighbor \cite{douze2024faiss}.
\begin{align}
    CD_{var} &= \frac{\sum_{i = 1}^{mk}(\rho_i-\overline{\rho})^2}{mk-1} \label{eqn:cdv}
\end{align}

\paragraph{Cluster Distribution (PD\textsubscript{EEE})} \quad
It may also be useful to quantify whether the data within each cluster (assigned to each centroid) are distributed irregularly. We compute the EEE (Equation \ref{eq:EEE}) for each centroid and report the combined average over all centroids and subspaces (Equation \ref{eqn:pdEEE}). For a uniform latent space, or locally uniform clusters, we expect the PD\textsubscript{EEE} to be low, and with more irregularly shaped clusters we expect the PD\textsubscript{EEE} to be higher.

\vspace{-0.2in}
\begin{align}
    PD_{EEE} &= \frac{\sum_{i=1}^{mk}EEE(x_i)}{mk} \label{eqn:pdEEE}
\end{align}

\subsection{Non-Standard Models}\label{sec:alt_models}
We also applied our measures to several models from the literature that had non-linguistic or otherwise challenging alterations made to the training protocol before fine-tuning on the standard (English) GLUE benchmarking tasks. If our measures can explain the surprising GLUE performance of these models, this would further support the notion that there were something about the data organization in the latent space that contributed directly to learning downstream tasks. The models that we included in our experiment, which we term ``non-standard'' models for simplicity, are briefly described below.

From \citet{zhang2021general}:
\begin{itemize}
\itemsep -0.3em
\item \textbf{Chinese}: a standard pre-trained Chinese BERT-base model with the input embeddings swapped out for the pre-trained English BERT embeddings;
\item \textbf{German} - a standard pre-trained German BERT-base model with the input embeddings swapped out for the pre-trained English BERT embeddings;
\item \textbf{Shuffle}: a standard pre-trained English BERT-base model with the indices of the input embeddings shuffled;
\item \textbf{Untrained}: a randomly initialized BERT-base model with the pre-trained English BERT embeddings. Note that the BERT-base model is initialized using a $N(0,0.02)$  distribution, so the results of this model are expected to differ from a model with 100\% weight perturbation pulled from a $N(0,1)$ distribution.
\end{itemize}
From \citet{alajrami2022does}:
\begin{itemize}
\itemsep -0.2em
\item \textbf{MLM}: A RoBERTa model pre-trained with the traditional masked language modeling task alone;
\item \textbf{First Char}: A RoBERTa model pre-trained with the task of predicting the first character of the masked tokens;
\item \textbf{ASCII} : A RoBERTa model pre-trained with the task of predicting the sum of the ASCII characters of the masked tokens;
\item \textbf{Random}: A RoBERTa model pre-trained with the task of predicting a random number.
\end{itemize}
From \citet{sinha2021masked}:
\begin{itemize}
\itemsep -0.2em
\item \textbf{RoBERTa}: A standard pre-trained RoBERTa model using the same training regimen (datasets, length of training, etc.) as the other models in the paper;
\item \textbf{Sequence}: A RoBERTa model trained with words shuffled within each input sequence;
\item \textbf{Corpus}: A RoBERTa model trained with sequences sampled from the frequency distribution of the entire corpus (analogous to shuffling words within the whole corpus).
\end{itemize}

\section{Results}

\subsection{Cell Density Measures} \label{sec:patch_results}

\begin{figure}
\begin{center}
\centerline{\includegraphics[width = 0.7\columnwidth]{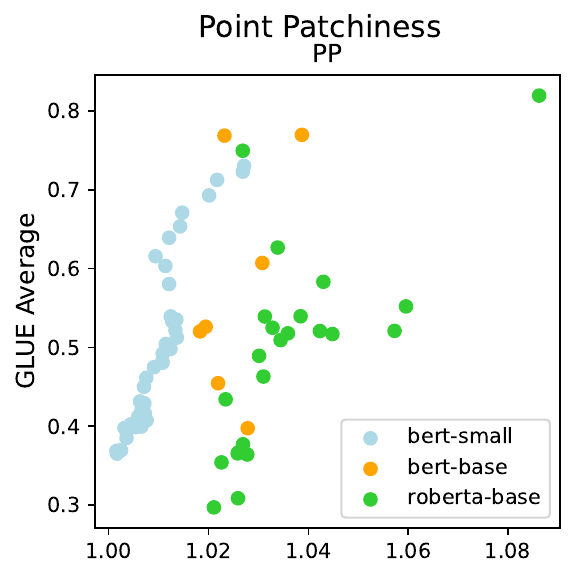}}
\vskip -0.1in
\caption{Point Patchiness (PP) shows a positive linear correlation with Average GLUE score. Pearson's $r = 0.9$ for perturbed BERT-small models and $r=0.64$ for RoBERTa models. }
\label{fig:aq_patch}
\end{center}
\vskip -0.3in
\end{figure}

Figure \ref{fig:aq_patch} shows the strong linear relationship between GLUE scores and a density measure (patchiness) on BERT-small models (Pearson's $r=0.902$)\footnote{We produced triplicate perturbed models for all $\alpha$-levels of the BERT-small models, and on RoBERTa models with $\alpha < 0.1$, and report the median value on our charts to ease interpretation.}. These results support our intuition that a less uniform data distribution is more conducive to learning downstream tasks, since we expect the variation in cell density (patchiness) to increase in a less isotropic data distribution.

A weaker linear relationship is also apparent with the RoBERTa models ($r=0.640$), with reduction in predictive value particularly at higher GLUE (lower $\alpha$) values. This plot also shows that both the RoBERTa and BERT-base model results are consistently shifted to the right (with higher PP values) when compared to the BERT-small results. This shift could be caused by any number of differences between the models, including number of parameters, length of training, size of training dataset, and, for RoBERTa, several differences in the pre-training tasks.

\begin{figure}
\begin{center}
\centerline{\includegraphics[width = \columnwidth]{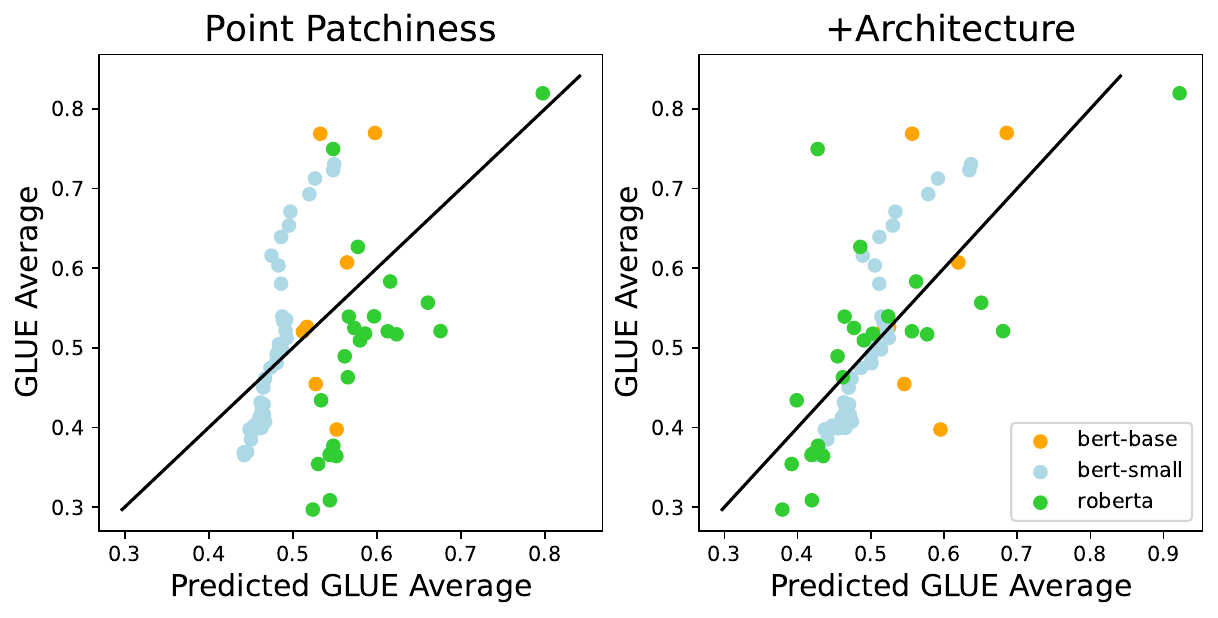}}
\vskip -0.1in
\caption{Residual plots for Point Patchiness linear regression model with model architecture variable improved the overall data fit ($r^2=0.3$ for the PP model, $r^2=0.5$ for the +Architecture Model) .}
\label{fig:aq_patch_lm}
\end{center}
\vskip -0.4in
\end{figure}

The predictive utility of the patchiness index is shown in Figure \ref{fig:aq_patch_lm}. Including the model architecture (BERT-small vs. BERT-base vs. RoBERTa) as a variable in a simple linear regression model yields an overall model fit of $r^2 =0.5$. 

\begin{figure}
\begin{center}
\centerline{\includegraphics[width = 0.7\columnwidth]{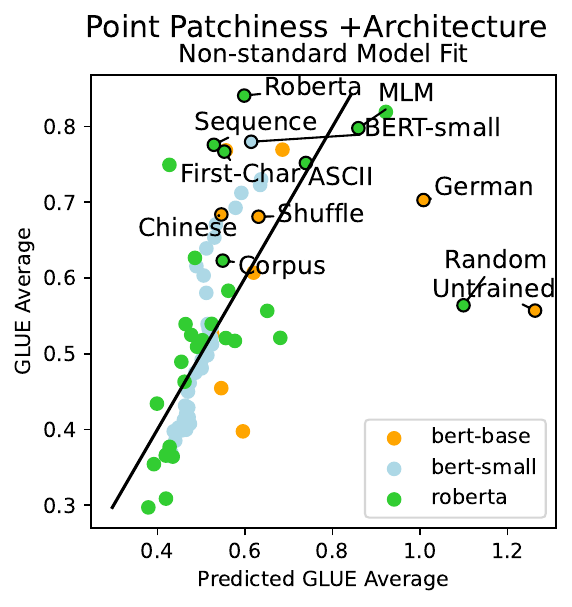}}
\vskip -0.1in
\caption{Point Patchiness linear regression model is predictive of some, but not all non-standard models from the literature.}
\label{fig:aq_patch_alt}
\end{center}
\vskip -0.4in
\end{figure}

The linear relationship seen for our perturbed model data is less clear for the non-standard models described in Section \ref{sec:alt_models}. Figure \ref{fig:aq_patch_alt} includes these non-standard models in the linear regression plot. Although the model seems predictive for many of the models when we include a model architecture variable, there are still a few models that are clear outliers and remain unexplained by this measure alone.

These results demonstrate that, while there is a suggestive relationship between the patchiness measure and GLUE score, it is not fully explanatory. It is certainly possible that factors other than the latent space data organization  contribute to the performance of these unexplained non-standard models, but it is also likely that our measures, which merely approximate some geometric characteristics of the space, do not fully capture a relationship that does exist.

\subsection{Other Measures}\label{sec:other_results}

\begin{figure*}
\begin{center}
\centerline{\includegraphics[width = 0.9\textwidth]{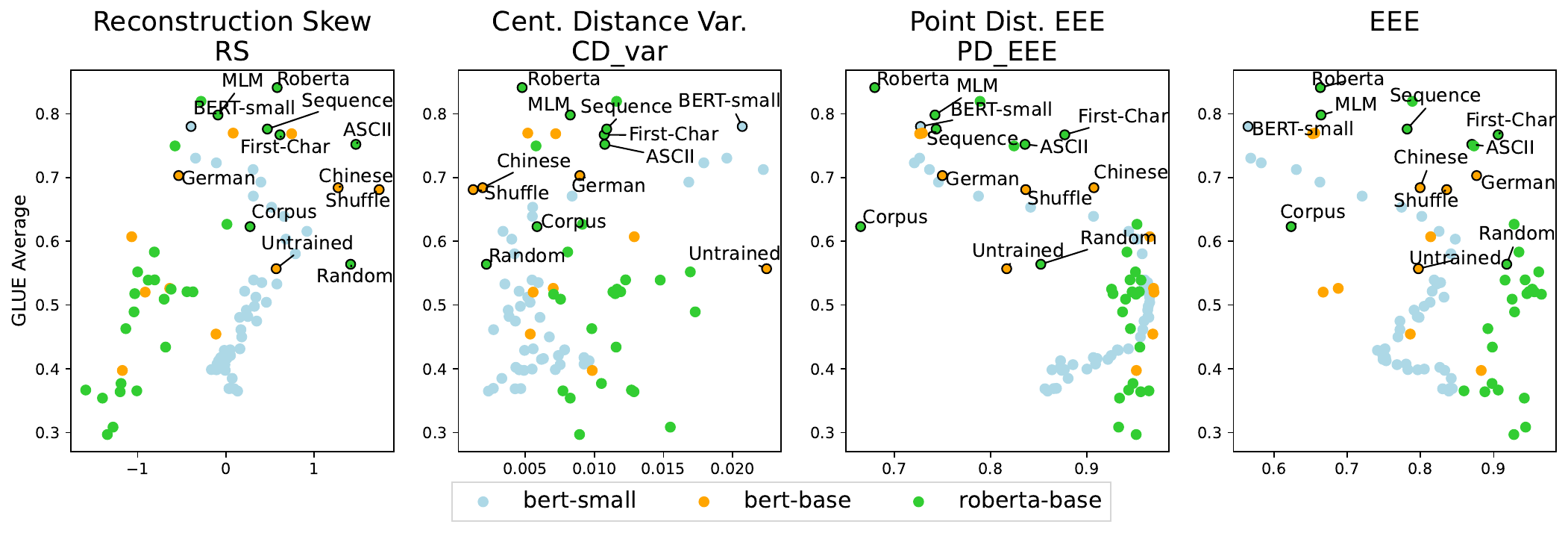}}
\vskip -0.1in
\caption{Other measures show non-linear relationships that are non-monotonic with the average GLUE score.}
\label{fig:other_metrics}
\end{center}
\vskip -0.2in
\end{figure*}

While the cell density measures are the most promising candidates for linear prediction of GLUE performance, the results in Figure \ref{fig:other_metrics} show that other measures have a more complex relationship with the average GLUE score. When looking at the BERT-small results in particular, these relationships are clearly non-linear, and often non-monotonic in the measure value. Several of our measures result in the ``zig-zag'' shape seen clearly on the EEE plot, and several others have more of the ``hook'' shape seen on the PD-EEE plot. The full set of measure plots can be found in Appendix \ref{app:full_results}.

\begin{table}
    \vskip 0.1in
    \caption{Pearson's correlation values between select measures and average GLUE scores for BERT-small (n=125) and RoBERTa (n=36) models with perturbed weights; MSE of linear regression (LM) with model architecture type when applied to non-standard models described in Section \ref{sec:alt_models}. Values with MSE magnitude $\geq 0.6$ are bolded}
        \vskip -0.1in
	\label{tbl:correlation}
        \begin{center}
        \begin{small}
        \begin{sc}
	\begin{tabular}{|l|c|c|c|}\hline
           \multirow{2}{*}{}& \multicolumn{2}{c|}{Pearson's $r$} & LM \Tstrut\\\cline{2-3}
           & BERT-small & RoBERTa & MSE \Tstrut\\\hline
           PP & \textbf{0.902} & \textbf{0.640} & 0.089 \Tstrut\\
           RS & 0.461 & -0.574 & 0.039 \\
           CD\textsubscript{var} & 0.537 & -0.292 & 0.053 \\
           PD\textsubscript{EEE} & -0.429 & \textbf{-0.764} & 0.017 \\
           EEE & -0.557 & -0.535 & 0.054 \\\hline
           
	\end{tabular}
        \end{sc}
        \end{small}
        \end{center}
        \vskip -0.15in
\end{table}

\begin{figure}
\vspace{-0.2in}
\begin{center}
\centerline{\includegraphics[width = 0.7\columnwidth]{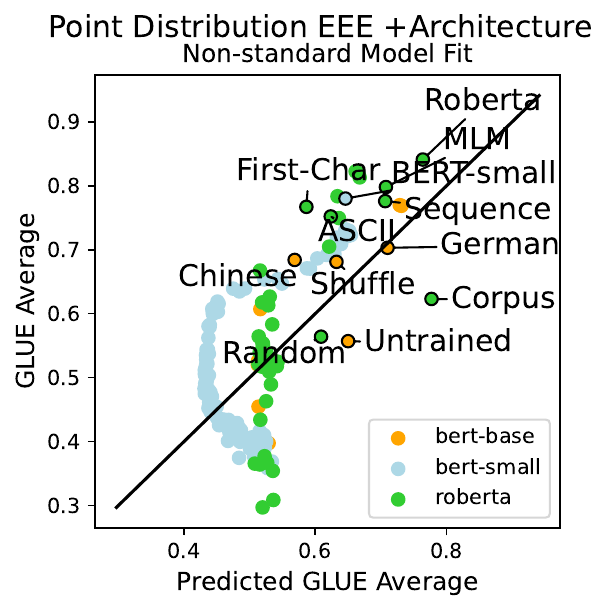}}
\vskip -0.1in
\caption{Although the PD\textsubscript{EEE} linear regression has a very low MSE for the non-standard models from the literature, the residuals plot shows that a linear model is not a good fit for the distribution overall.}
\label{fig:pdEEE}
\end{center}
\vskip -0.4in
\end{figure}

Given these non-linear relationships, the results are surprising when we apply linear regression models of the measures to predict the GLUE scores of the non-standard models described in Section \ref{sec:alt_models}. In the rightmost column of Table \ref{tbl:correlation}, we see that the PD\textsubscript{EEE} model has the lowest MSE, even compared to the PP model after adding the model architecture information. However, the curved shape on the residual plot for this model (Figure \ref{fig:pdEEE}) illustrates that, even though the non-standard models are fairly well-predicted by the model, a linear model is not a good fit for the data more generally. 

We explored the utility of linear regression models with higher order variables, combinations of different measures, and interactions between measures as well. While some of these models showed minimal improvement over the simple models discussed above, we determined that the loss of interpretability in these more complex models outweighed the improvement to model fit.

Although these relationships are interesting and may be informative about the inner workings of the transformer models themselves, the non-monotonicity of these measures in relation to the GLUE average makes them poor candidates for predicting downstream task performance. For example, we cannot say that lowering the EEE or PD\textsubscript{EEE} of a latent space will necessarily improve benchmarking performance. We offer some discussion of these non-linear relationships in Appendix \ref{app:zigzag}, but leave any further exploration of these results to future work.

\section{Conclusion}
In this work, we address the question of whether there is some measurable geometric characteristic to the latent space of transformer models that could be used to predict performance on downstream tasks, specifically the GLUE benchmarking tasks. We produced models with gradually decreasing GLUE scores by transitioning the weights of pre-trained language models to random noise, and then we applied a set of measures chosen to approximately capture the amount of data spread and separability present in the resulting models' latent spaces. 

In examining the relationship between these measures and GLUE benchmarking performance, we identified a quantized cell density measure, based on the patchiness index~\cite{wade201850}, that has a strong linear relationship to the GLUE score. We also considered whether any of these measures could be used to predict the unexpected GLUE performance found in a body of work exploring non-standard training tasks and model manipulations. Although the evidence was less clear, we once again found that the patchiness measure showed promise as a predictive measure for the GLUE score of most of the non-standard models that we examined.

This research is potentially relevant to the growing body of work attempting to reduce the resource requirements for pre-training large language models. If a measure of separability within a model's utilized latent space could be used to initialize a transformer model before any linguistic training begins, it is possible that the pre-training requirements could be greatly reduced without sacrificing performance on downstream tasks. 

Our experiments also uncovered interesting non-linear relationships between many of the measures that we examined and average GLUE score. Further exploration of these non-monotonic patterns could shed light on the inner workings of transformer models, contributing to work on increasing the general interpretability of transformer models.

\section{Limitations}

Our experiments were limited by two major difficulties: (1) The practically infinite nature of a contextual model's latent space, and (2) the lack of established measures for describing high dimensional data distributions. To overcome these obstacles, we had to accept approximate solutions to both.

Our measure calculations are based on a sample of the latent space that we showed to be representative of a sizable corpus, but is still necessarily an imperfect approximation of the theoretical complete latent space. Additionally, and perhaps more relevant to our findings, the contextual nature of the latent spaces required manipulation of the transformer model itself to approximate a gradual change in the geometry of the latent space. As briefly discussed in Appendix \ref{app:zigzag}, some of our results clearly indicate that adding noise to the model weights is not the same as adding noise to the latent space representations, which would perhaps provide a clearer signal for our intended experiment.

While the data spread measures that we examined are fairly well understood, our use of quantization algorithms to approximate some measure of the clustered nature of the data introduces a bit of a ``black box'' into the measure calculations. The algorithms are designed to effectively segment the data for optimal reconstruction, and it is unclear how this goal interacts with some of the measures that we apply. For example, the quantized cell density measure (for which we found the most compelling results) is based on the distribution of the number of points assigned to each centroid in the quantized space. However, it may be the case that the quantization algorithms themselves are in direct conflict with this measure, specifically attempting to adjust the location of the centroids and subspaces to even out the distribution of points assigned to each cluster.
Also, the number of quantized components may not adequately describe the complexity of space utilization -- imposition of too many or too few components may play a role in some of the observed non-monotonic relationship between density measure and GLUE score. We do not believe these to be issues for our experimental results, but it does hinder our understanding of how our measures are related to the latent space geometry.

Additionally, we remain unsatisfied that any of our measures quite captured the concept of separability that we intended, somehow quantifying the extent to which data is located in ``nooks and crannies'' in the latent space. We point this out not to discredit our work and results, but to again highlight the fact that all of these measures are approximations of complex geometric relationships. It is certainly possible that a different measure exists that would better capture the relationship between GLUE performance and latent space organization, and it is perhaps equally possible that the relationship is too complex to be fully approximated by a single measure.

Finally, with the future goal of using this paper's results to reduce pre-training requirements for transformer models, we come up against the major issue that existing quantization methods are non-differentiable. This means that none of the quantization-based measures explored here can be used as an additional or initial loss function. However, we do intend to explore the possibility of using differentiable k-means algorithms in place of the quantization methods used here to address this problem in future work.

\section{Acknowledgements}
We thank Katy McKinney-Bock, Conner Copeland, and Daniel Olson for enlightening conversations during the development of this project. 
Computational analysis was performed on Jetstream2 resources at Indiana University through allocation  BIO220047 from the NSF ACCESS program, which is supported by OAC Grants Nos. 2138259, 2138286, 2138307, 2137603, and 2138296.

\bibliography{biblio}

\newpage

\appendix

\section{Other Measures}\label{app:other_metrics}
\begin{figure}[h]
\begin{center}
\centerline{\includegraphics[width = \columnwidth]{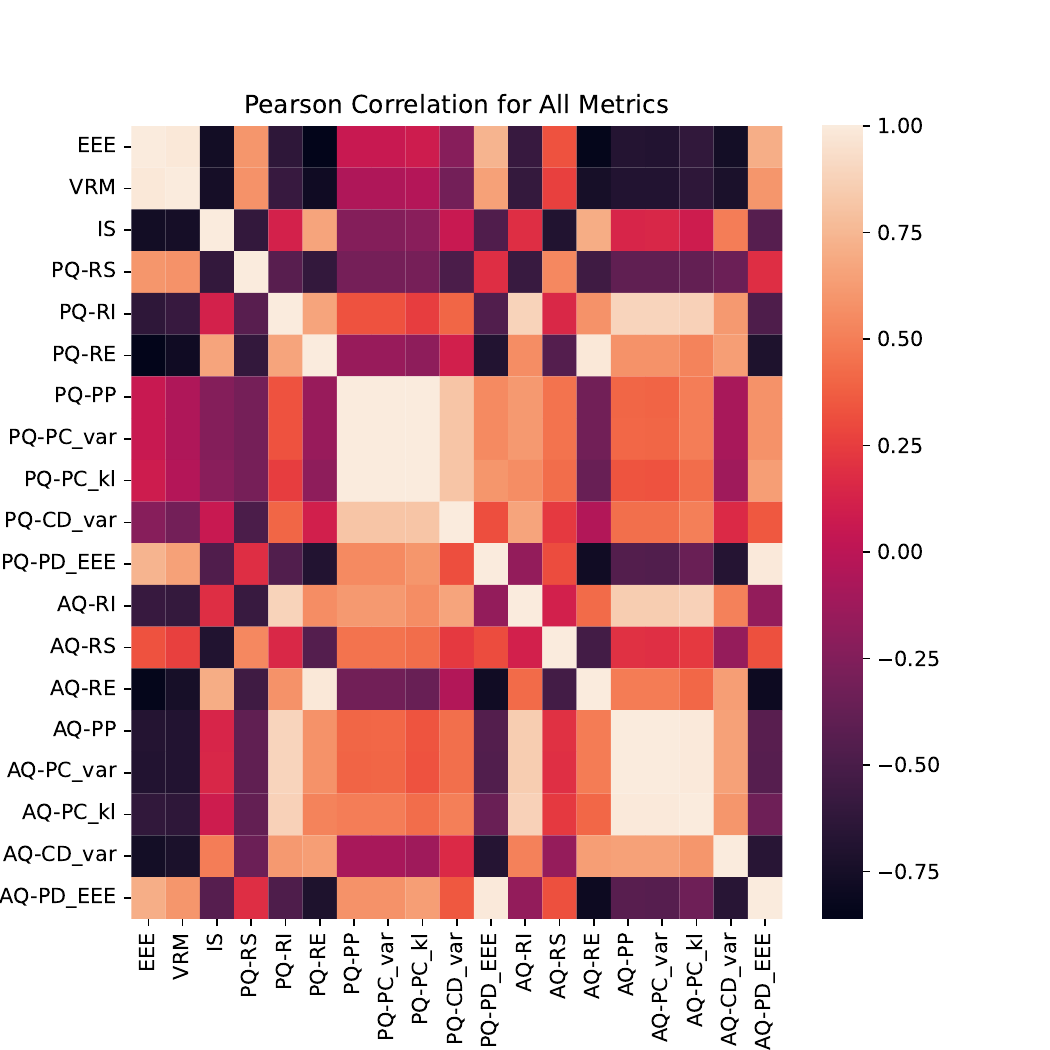}}
\vskip -0.1in
\caption{Pearson's Correlation Matrix for all measures.}
\label{fig:full_corr}
\end{center}
\vskip -0.4in
\end{figure}

Many of the measures that we included in our experiments were highly correlated (see Figure \ref{fig:full_corr}). This is not an unexpected result, as we intentionally chose measures that would be conceptually redundant in case one captured the distributional characteristic better than others. The detailed descriptions for the measures not included in the main paper can be found below.

\subsection{Data Spread}
In addition to the EEE measure, we included two other measures of overall data spread in our experiments, described below. VRM, like EEE, is designed to measure how evenly data are spread in all directions and along each dimension, while IsoScore is only designed to measure isotropy, or the equal use of all directions. EEE and VRM are very highly correlated, with IsoScore only slightly less highly correlated.

\paragraph{Vasicek Ratio Mean Squared Error (VRM)} \quad
Following \citet{marbut2023reliable}, we compute a measure based on the Vasicek entropy approximation as presented in Equation \ref{eq:vasicek}. This approximation considers pairs of ordered points that are separated by a fixed interval, $m$, where we would expect the distances between these points to be equal in an evenly spread distribution, and highly variant (entropic) in an uneven distribution. $H_{vas}$ is then used in a ratio with the theoretical value for a reference normal distribution, and the mean squared error (MSE) is computed to produce a multi-dimensional measure (Equation~\ref{eq:vac_multi}). We expect the VRM of a distribution to decrease as data become more evenly spread.
\begin{align}
H_{vas} &= \frac{1}{n}\sum_{i=1}^{n}\log \frac{n}{2m}(x_{i+m}-x_{i-m}) \label{eq:vasicek} \\
VRM&= \frac{1}{d}\sum\limits_{i=1}^{d}(1-\frac{H_{vas}}{\ln(\sqrt{2\pi e}\sigma^{2})})^2 \label{eq:vac_multi}
\end{align}

\paragraph{IsoScore (IS)} \quad
IsoScore, from \citet{rudman2021isoscore}, approximates the proportion of dimensions that are isotropically used as shown in Equation \ref{eq:isoscore}. This is based on the \emph{isotropy defect} (Equation \ref{eq:defect}), which is essentially the distance between the normalized covariance matrix of a latent space, $\hat{\sum}_D$, and the identity matrix. We expect the IS of a distribution to increase as data becomes more isotropically spread.
\begin{align}
\delta(X) &= \frac{\Vert\hat{\sum}_{D}-1\Vert}{\sqrt{2(n-\sqrt{n})}}\label{eq:defect}\\
IS &= \frac{(n-\delta(X)^2(n-\sqrt{n}))^2-n}{n(n-1)}\label{eq:isoscore}
\end{align}

\subsection{Quantized Cell Density}
In addition to the Point Patchiness measure, we included two other measures of quantized cell density in our experiments. All three of these measures were highly correlated, so we chose to only include the Point Patchiness measure in our main discussion.

\paragraph{Point Count Distribution (PC\textsubscript{var} \& PC\textsubscript{kl})} \quad  As alternative measures of cell density, we also consider the variance (Equation \ref{eqn:pcvar}) and the KL-divergence compared to a uniform distribution (Equation \ref{eqn:pckl}) of the odds-ratio of points assigned to the $k$ centroids (Equation \ref{eqn:odds}), as averaged over the $m$ subspaces.
\begin{align}
    O_i &= (k-1)\frac{c_i}{1-c_i}\label{eqn:odds} \\
    PC_{var} &= \frac{\sum_{i=1}^{mk}(O_i-\overline{O})^2}{m(k-1)} \label{eqn:pcvar} \\
    PC_{kl} &=  \frac{\sum_{i=1}^{mk}p(O_{i})\log\frac{p(O_i)}{q(O_i)}}{m}  \label{eqn:pckl}
\end{align}
Here, $c_i$ is the normalized count of points assigned to centroid $i$, $\overline{O}$ is the mean of the odds ratio distribution, and $p(O)$ and $q(O)$ are the observed and reference uniform odds ratio distributions, respectively.

\subsection{Quantization Reconstruction}
In addition to the Reconstruction Skew measure, we included two other reconstruction-based measures in our experiments. Although these were not highly correlated with each other, they were correlated with other measures included in the main discussion.

\paragraph{Reconstruction Error (RE)} \quad
We consider Reconstruction Error as a potential measure of separability. We expect that a distribution that is already organized in clusters would result in quantized (compressed) representations that are close to the original data, resulting in lower error than a distribution that is not as easily clustered. We calculate a simple element-wise sum squared error between the original data, $x$, and the reconstructed data, $x'$, as shown in Equation \ref{eqn:rec_error}. 
\begin{align}
    RE &= \frac{\sum(x_i-x'_i)^2}{\sum x_{i}^{2}}\label{eqn:rec_error}
\end{align}

\paragraph{Reconstruction IQR (RI)} \quad
We also consider the width of the inner quartile range (IQR) for the distribution of normalized error magnitudes (Equation \ref{eqn:error_mag}), where the IQR is defined as the range between the 75th and 25th percentiles of the distribution. For a more uniform distribution we expect little variation in the error magnitudes, resulting in a very small IQR. If the data were distributed irregularly, we would expect some points to be very close to their reconstructed counterparts, and some to be relatively far, causing a larger IQR.

\section{Non-Linear Relationship Discussion}\label{app:zigzag}

While it is not central to our work in this paper, the non-linear and non-monotonic relationships apparent between many of our proposed measures and average GLUE score were too intriguing not to address. We leave some ideas and light research notes here in hopes that it may spark further investigation.

One finding of note that comes out of the zig-zag pattern seen in many of our measures is that the addition of noise to the weights of a pre-trained BERT-small model does not equate to the addition of noise in the model's latent space. This is particularly evident in the results for the data spread measures, which are specifically designed to measure how evenly the data is spread. Results for all three data spread measures are reversed from our expectation, with the measures generally showing less even spread as noise is added to the model weights (with the exception of the central ``zig'' section of the plots). This also supports the argument against the findings that increasing isotropy improves downstream task performance, as discussed in \citet{mickus2024isotropy} and \citet{rudman2023stable}.

In our experiment, we gradually add the same random noise (using triplicate random seeds for each $\alpha$-level on the BERT-small models, and for small $\alpha$-levels in the RoBERTa models) to the same pre-trained model weights. \citet{zhou2019deconstructing} explore the idea of ``lottery tickets'' within trained transformer models, wherein a subset of the model weights can be used to reproduce (or even improve on) the full model's performance. One possible explanation for the non-monotonic behavior of the measure results is that in gradually adding noise to all of the pre-trained weights, we momentarily isolate or accentuate the influence of weights on one or more of these subnetworks along the way to actual random noise.

It has also been shown that the weights in different layers of transformer models are affected unevenly during training \cite{merchant2020happens, valeriani2024geometry}. Once again, we treat all weights in the model equally during perturbation, so it may be that the signals from certain layers take more or less noise to fully ablate. It is plausible that this could cause a momentary reversal in our latent space measures, resulting in the ``zig-zag'' shape observed for several of the measures.

\section{Full Results}\label{app:full_results}

The table and figures on the following pages provide a view into the predictive nature of the expanded set of measures introduced in this Appendix.

\begin{table*}
    \vskip -0.1in
    \caption{Pearson's correlation values between select measures and average GLUE scores for BERT-small (n=125) and RoBERTa (n=36) models with perturbed weights; MSE of linear regression with model architecture type when applied to non-standard models described in Section \ref{sec:alt_models}.}
        \vskip -0.1in
	\label{tbl:full_correlation}
        \begin{center}
        \begin{small}
        \begin{sc}
	\begin{tabular}{|cl|c|c|c|}\hline
           \multirow{2}{*}{}&\multirow{2}{*}{}& \multicolumn{2}{c|}{Pearson's $r$} & LM \Tstrut\\\cline{3-5}
           & & BERT-small & RoBERTa & MSE \Tstrut\\\hline
           \multirow{8}{*}{\rotatebox[origin = c]{90}{\parbox{1.5cm}{Additive \\ Quantizer}}} & \multicolumn{1}{|l|}{PP} & \textbf{0.902} & \textbf{0.640} & 0.089 \Tstrut\\
           & \multicolumn{1}{|l|}{PC\textsubscript{var}} & \textbf{0.901} & \textbf{0.652} & 0.082\\
           & \multicolumn{1}{|l|}{PC\textsubscript{kl}} & \textbf{0.890} & 0.228 & 0.058 \\
           & \multicolumn{1}{|l|}{RS} & 0.461 & -0.574 & 0.039 \\
           & \multicolumn{1}{|l|}{RE} & 0.334 & 0.227 & 0.023 \\
           & \multicolumn{1}{|l|}{RI} & \textbf{0.695} & -0.051 & 0.116\\
           & \multicolumn{1}{|l|}{CD\textsubscript{var}} & 0.537 & -0.292 & 0.053 \\
           & \multicolumn{1}{|l|}{PD\textsubscript{EEE}} & -0.429 & \textbf{-0.764} & 0.017 \\\hline
           \multirow{8}{*}{\rotatebox[origin = c]{90}{\parbox{1.5cm}{Product \\ Quantizer}}} & \multicolumn{1}{|l|}{PP} & 0.326 & -0.418 & 0.029 \Tstrut\\
           & \multicolumn{1}{|l|}{PC\textsubscript{var}} & 0.327 & -0.415 & 0.029\\
           & \multicolumn{1}{|l|}{PC\textsubscript{kl}} & 0.260 & 0.496 & 0.032 \\
           & \multicolumn{1}{|l|}{RS} & -0.123 & -0.184 & 0.034 \\
           & \multicolumn{1}{|l|}{RE} & 0.420 & 0.001 & 0.024 \\
           & \multicolumn{1}{|l|}{RI} & \textbf{0.808} & 0.514 & 0.082\\
           & \multicolumn{1}{|l|}{CD\textsubscript{var}} & 0.342 & -0.422 & 0.037 \\
           & \multicolumn{1}{|l|}{PD\textsubscript{EEE}} & -0.456 & \textbf{-0.705} & 0.026 \\\hline
           \multirow{3}{*}{\rotatebox[origin = c]{90}{\parbox{0.9 cm}{Data \\ Spread}}} & \multicolumn{1}{|l|}{EEE} & -0.557 & -0.535 & 0.054 \Tstrut\\
           & \multicolumn{1}{|l|}{VRM} & -0.573 & -0.534 & 0.016\\
           & \multicolumn{1}{|l|}{IS} & -0.078 & -0.398 & 0.035\\\hline           
	\end{tabular}
        \end{sc}
        \end{small}
        \end{center}
        \vskip -0.15in
\end{table*}

\begin{figure*}
\begin{center}
\centerline{\includegraphics[width = 0.9\textwidth]{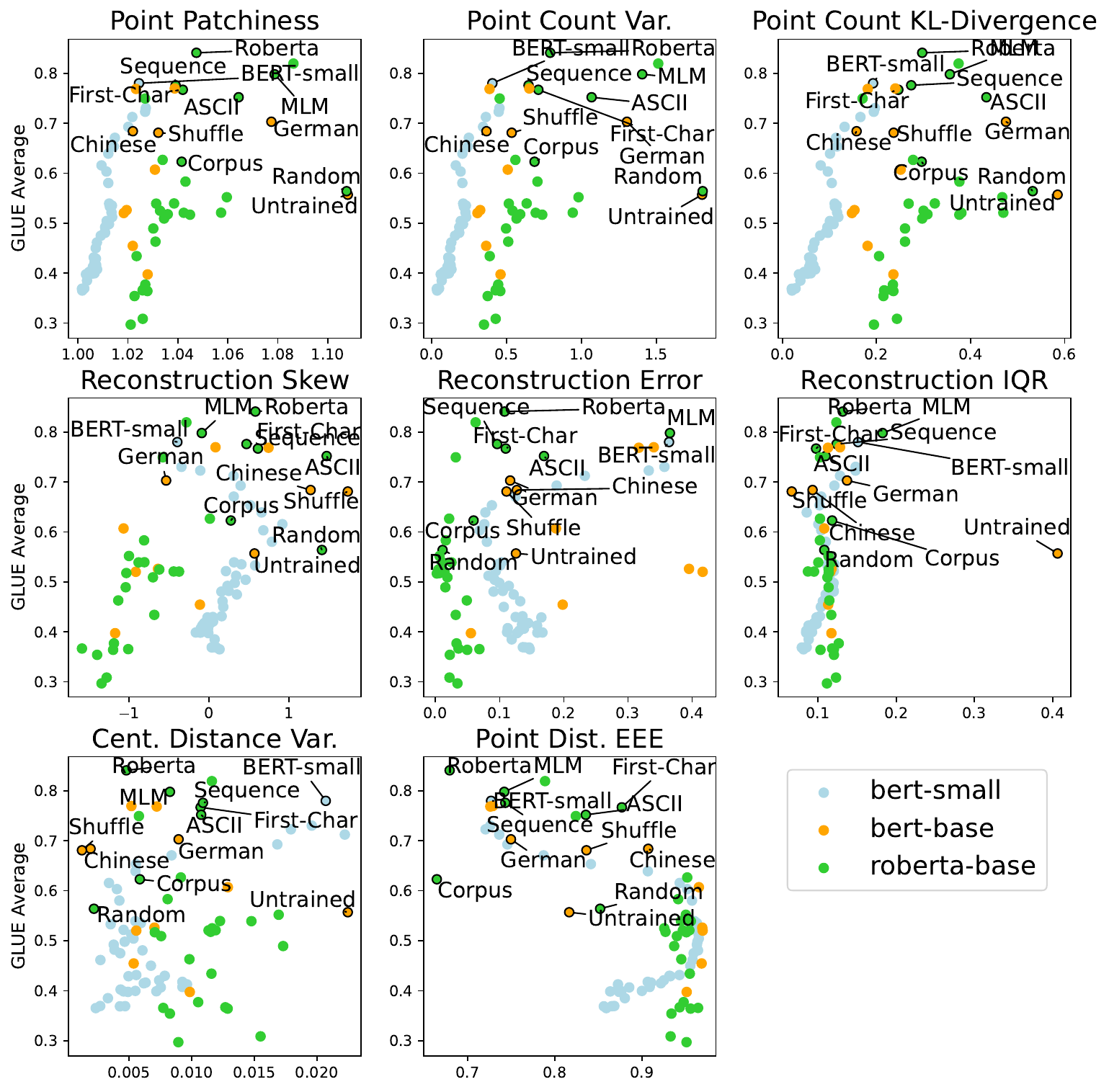}}
\vskip -0.1in
\caption{Average GLUE relationships for all Additive Quantizer Measures.}
\label{fig:full_aq}
\end{center}
\vskip -0.4in
\end{figure*}

\begin{figure*}
\begin{center}
\centerline{\includegraphics[width = 0.9\textwidth]{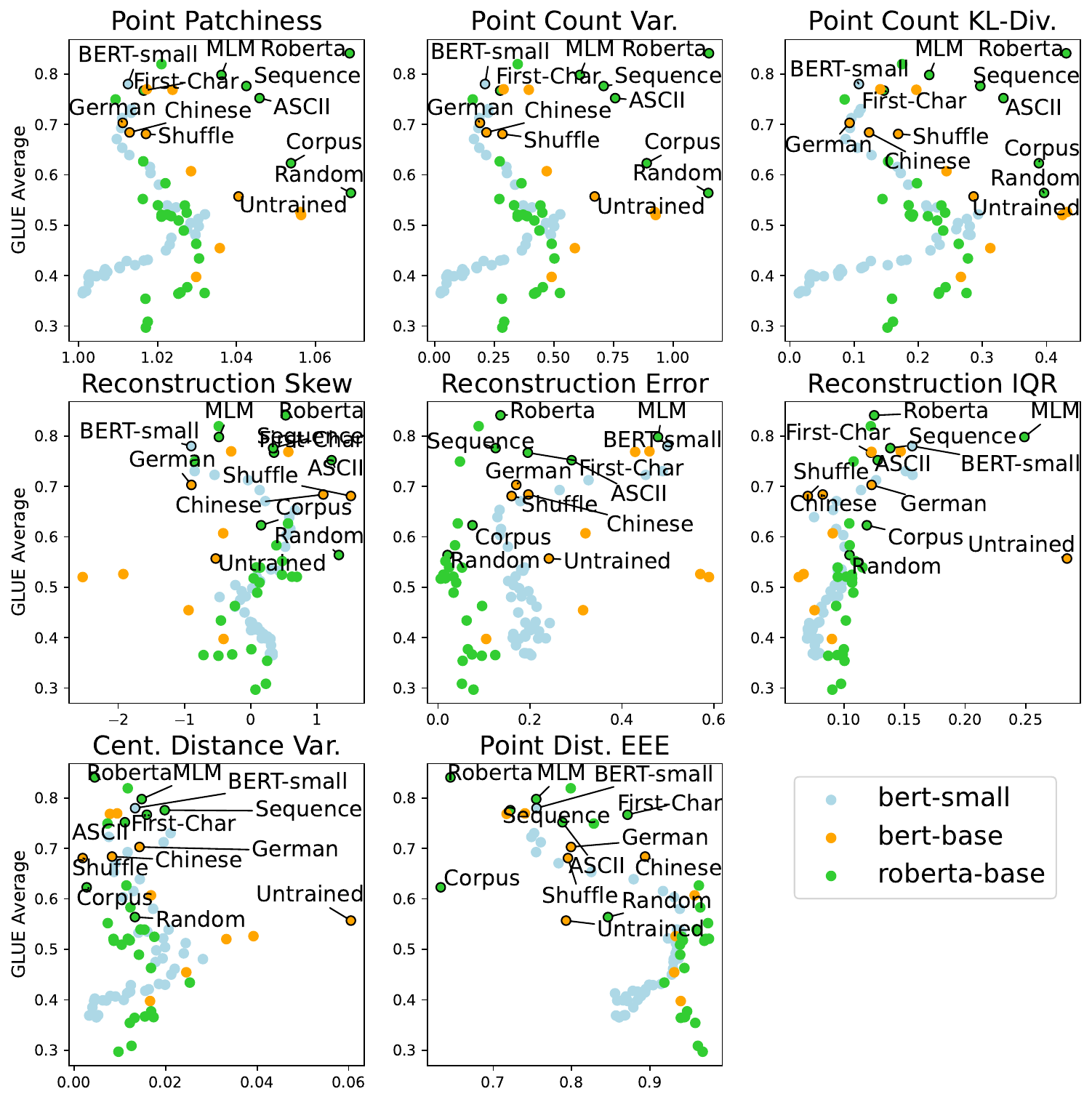}}
\vskip -0.1in
\caption{Average GLUE relationships for all Product Quantizer Measures.}
\label{fig:full_pq}
\end{center}
\vskip -0.4in
\end{figure*}

\begin{figure*}
\begin{center}
\centerline{\includegraphics[width = 0.9\textwidth]{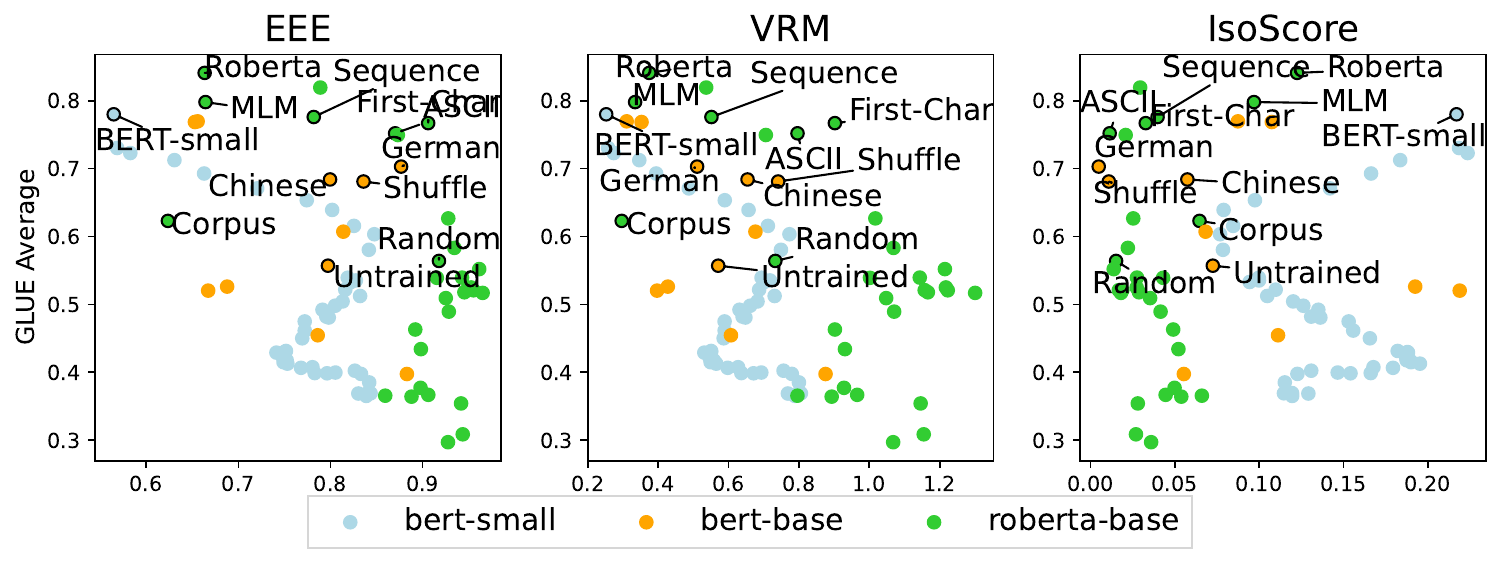}}
\vskip -0.1in
\caption{Average GLUE relationships for all Data Spread Measures.}
\label{fig:full_spread}
\end{center}
\vskip -0.4in
\end{figure*}

\end{document}